\documentclass[9pt, conference]{IEEEtran}
\IEEEoverridecommandlockouts
\usepackage{cite}
\usepackage{amsmath,amssymb,amsfonts}
\usepackage{algorithmic}
\usepackage{graphicx}
\usepackage{textcomp}
\usepackage{booktabs}
\usepackage{multirow}
\usepackage{tcolorbox}
\usepackage{xcolor,colortbl}
\usepackage{hyperref}
\usepackage{cleveref}
\newtcolorbox{todo}[1][]{colback=red!5!white, colframe=red!75!black, title=TODO, #1}

\newtcolorbox{notice}[1][]{colback=yellow!5!white, colframe=yellow!75!black, title=NOTICE, #1}

\definecolor{lightergray}{gray}{0.92}

\def\BibTeX{{\rm B\kern-.05em{\sc i\kern-.025em b}\kern-.08em
    T\kern-.1667em\lower.7ex\hbox{E}\kern-.125emX}}

\newcommand{\verticalrow}[2]{\parbox[t]{2mm}{\multirow{#1}{*}{\rotatebox[origin=c]{90}{#2}}}}

\begin{document}

\title{Effective Context Modeling Framework for Emotion Recognition in Conversations
\thanks{
$^*$ These authors contributed equally to this work.
}
\thanks{
$^\dagger$ Correspondence to {\color{blue}thy@uab.edu}
}
}


\author{\IEEEauthorblockN{Cuong Tran Van$^{*,1}$\qquad Thanh V. T. Tran$^{*,1}$\qquad Van Nguyen$^1$\qquad Truong Son Hy$^{1,2,\dagger}$}
\IEEEauthorblockA{$^1$\textit{FPT Software AI Center, Hanoi, Vietnam} \\
$^2$\textit{Department of Computer Science, The University of Alabama at Birmingham, United States}}
}

\maketitle

\begin{abstract}
Emotion Recognition in Conversations (ERC) facilitates a deeper understanding of the emotions conveyed by speakers in each utterance within a conversation. Recently, Graph Neural Networks (GNNs) have demonstrated their strengths in capturing data relationships, particularly in contextual information modeling and multimodal fusion. However, existing methods often struggle to fully capture the complex interactions between multiple modalities and conversational context, limiting their expressiveness. To overcome these limitations, we propose ConxGNN, a novel GNN-based framework designed to capture contextual information in conversations. ConxGNN features two key parallel modules: a multi-scale heterogeneous graph that captures the diverse effects of utterances on emotional changes, and a hypergraph that models the multivariate relationships among modalities and utterances. The outputs from these modules are integrated into a fusion layer, where a cross-modal attention mechanism is applied to produce a contextually enriched representation. Additionally, ConxGNN tackles the challenge of recognizing minority or semantically similar emotion classes by incorporating a re-weighting scheme into the loss functions. Experimental results on the IEMOCAP and MELD benchmark datasets demonstrate the effectiveness of our method, achieving state-of-the-art performance compared to previous baselines.
\end{abstract}

\section{Introduction}

Emotion Recognition in Conversations (ERC) has gained significant attention as a research field for its broad practical applications. Traditional ERC approaches primarily focus on classifying emotions within individual utterances using conversational text \cite{hu-etal-2021-dialoguecrn,10446496}. Leveraging the continuous nature of utterances in a conversation, some ERC methods model both the semantic features of utterances and the contextual information of conversations. Early approaches like ICON \cite{hazarika-etal-2018-icon}, CMN \cite{hazarika-etal-2018-conversational}, and DialogueRNN \cite{Majumder2019dialrnn} employ RNNs to model the conversation as a sequential flow of utterances. Meanwhile, Ada2I \cite{10.1145/3664647.3681648} tackled the challenge of modality imbalances and modality-fusion learning. However, these methods struggle to effectively balance long- and short-term dependencies for each utterance. In contrast, Graph Neural Networks (GNNs) have gained popularity due to their ability to efficiently aggregate information in conversational contexts. DialogueGCN \cite{ghosal-etal-2019-dialoguegcn} and RGAT \cite{ishiwatari-etal-2020-relation} employed GNNs to model inter-utterance and inter-speaker relationships. DAG \cite{shen-etal-2021-directed} leveraged the strengths of both traditional graph-based and recurrent neural networks. Recent advancements, such as CORECT \cite{nguyen-etal-2023-conversation} and M3GAT \cite{10.1145/3593583}, integrated modality-specific representations with cross-modal interactions to create more comprehensive models. Additionally, approaches like graph contrastive learning \cite{li-etal-2023-joyful,10.1145/3664647.3681633} and knowledge-aware GNNs \cite{10095097} further demonstrated the potential of GNNs to boost performance, setting a new benchmark for future ERC systems.

However, current GNN-based approaches still face limitations in fully capturing conversational context. First, they rely on a fixed window size to model contextual information for all utterances, overlooking the variability in emotional shifts across a dialogue. This fixed setting struggles to account for the different emotional influences of each utterance, as the range of emotional impact varies throughout conversations. Second, traditional GNNs assume pairwise relationships between nodes, while in ERC, the emotional tone of one utterance can influence multiple subsequent utterances, which cannot be effectively captured through pairwise connections alone. Third, the integration of fine-grained multimodal features into emotional state prediction has not been thoroughly explored, limiting the potential performance improvements. Finally, current state-of-the-art (SOTA) methods overlook the issue of class imbalance, where majority classes significantly outnumber minority classes. This imbalance results in suboptimal performance, particularly when predicting emotions from minority classes.

To address these issues, we introduce ConxGNN, a novel framework designed to fully capture contextual information in conversations. At its core, ConxGNN consists of two parallel components: the Inception Graph Module (IGM) and the Hypergraph Module (HM). Recognizing the varying impact of utterances across conversations, and inspired by the use of multiple filter sizes in \cite{Szegedy_2015_CVPR}, IGM is built with multiple branches, each using a different window size to model interaction distances between utterances, enabling multiscale context modeling. Simultaneously, we capture multivariate relationships within conversations by constructing a hypergraph neural network. The outputs of these two modules are then passed through an attention mechanism, where attention weights are learned to complement emotional information across modalities. Additionally, to mitigate class imbalance, we introduce a re-weighting term to the loss functions, including InfoNCE and cross-entropy loss. Experiments on two popular ERC datasets demonstrate that ConxGNN achieves best performance compared to SOTA methods. The contribution of this paper can be summarized as follows: \textbf{(1)} We propose ConxGNN, which effectively models both multi-scale and multivariate interactions among modalities and utterances; \textbf{(2)} We design an attention mechanism to integrate fine-grained features from both graph modules into a unified representation; \textbf{(3)} We address class imbalance with a re-weighting scheme in the loss functions; \textbf{(4)} We conduct experiments on the IEMOCAP and MELD datasets, demonstrating that our proposed method achieves SOTA performance across both benchmarks.




\section{Proposed Approach}\label{sec:proposed-approach}

\begin{figure*}[t]
    \centering
    \includegraphics[width=0.85\linewidth]{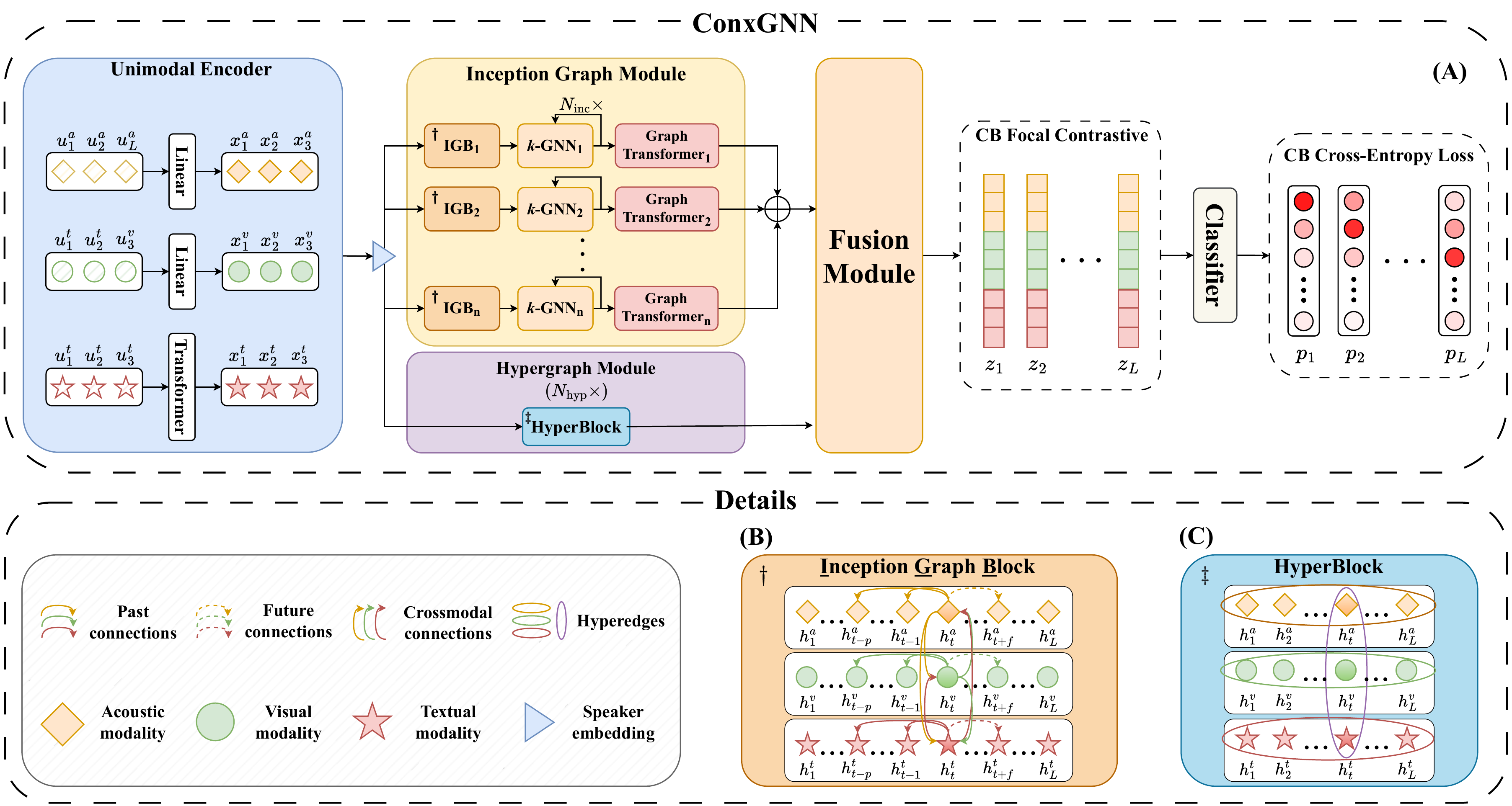}
    \caption{Detailed architecture of (A) the proposed ConxGNN, (B) Inception Graph Block, and (C) HyperBlock.}
    \label{fig:framework}
\end{figure*}

\subsection{Problem Formulation}

Given a conversation consisting of $L$ utterances $U = \{ u_1, u_2, \ldots, u_L \}$, where each utterance $u_i$ is spoken by speaker $s_i \in S$ and consists of multi-sensory data: textual $(\mathbf u_i^t)$, visual $(\mathbf u_i^v)$, and acoustic $(\mathbf u_i^a)$ modalities:
\begin{equation}
    u_i = \{ \mathbf u_i^t, \mathbf u_i^v, \mathbf u_i^a \},\quad i \in \{ 1, 2, \ldots, L \},
\end{equation}
in which $\mathbf u_i^\tau \in \mathbb R^{d_\tau}, \tau \in \{t,a,v\}$ with $d_\tau$ is the dimension size of raw modaltity features. The ERC task aims to predict the label for each utterance $u_i \in U$ from a set of $C$ predefined emotional labels $Y = \{ y_1, y_2,\ldots, y_C \}$.

Our proposed architecture is illustrated in \Cref{fig:framework}A. In general, ConxGNN contains five main components: a unimodal encoder, an inception graph module, a hypergraph module, a fusion module, and an emotion classifier.

\subsection{Unimodal Encoder}

Following \cite{nguyen-etal-2023-conversation}, we first capture the utterance-level features of each modality. Specifically, we utilize a Transformer encoder \cite{NIPS2017_3f5ee243} for textual modality and a fully-connected network for visual and acoustic modalities as follows:
\begin{align}
    \mathbf x_i^t &= \textbf{Transformer}(\mathbf u_i^t, \boldsymbol \theta_{\text{trans}}^t), \\
    \mathbf x_i^\tau &= \mathbf W^\tau \mathbf u_i^\tau + \mathbf b^\tau, \quad \tau \in \{a,v\},
\end{align}
where $\boldsymbol \theta_{\text{trans}}^t, \mathbf W^\tau \in \mathbb R^{d_h \times d_\tau}, \mathbf b^\tau \in \mathbb R^{d_h}$ are trainable parameters and $\mathbf x_i^t, \mathbf x_i^v, \mathbf x_i^a \in \mathbb R^{d_h}$. Additonally, considering the impact of speakers information in a conversation, we incorporate the embedding of speakers' identity and produce the respective latent representations $\mathbf s_i = \textbf{Embedding}(S)$, in which $\mathbf s_i \in \mathbb R^{d_h}$. We then add speaker embedding to obtain speaker- and context-aware unimodal representation $\mathbf h_i^\tau \in \mathbb R^{d_h}$ at the $i$-th conversation turn:
\begin{equation}
    \mathbf h_i^\tau = \mathbf s_i + \mathbf x_i^\tau, \quad \tau \in \{ t,a,v \}.
\end{equation}

\subsection{Inception Graph Module (IGM)}

\subsubsection{Graph Construction} 
We define $\mathcal G(\mathcal V_{\mathcal G}, \mathcal R_{\mathcal G}, \mathcal E_{\mathcal G})$ as the multimodal graph constructed from conversations. 

\textbf{Nodes.}$\quad$Each utterance is modeled as three distinct nodes, corresponding to the representations $\mathbf h_i^t, \mathbf h_i^v$ and $\mathbf h_i^a$, resulting in a total of $|\mathcal V| = 3L$ nodes.

\textbf{Relations.}$\quad$To capture both inter- and intra-dependencies among modalities, we define two types of relations: $\mathcal R_{\text{inter}}$ denotes the connections between the three modalities within the same utterance, while $\mathcal R_{\text{intra}}$ represents the connections between utterances of the same modality within a given time window. To capture this temporal aspect, we introduce a sliding window $[p, f]$ to control the number of past and future utterances that connected to the current node $\mathbf u_i^\tau$. Therefore, the two groups of relations can be expressed as follows:
\begin{align}
    \mathcal R_{\text{inter}} &= \big\{ (\mathbf h_i^\tau, \mathbf h_i^\nu) |\, \tau,\nu \in \{ t,a,v \}\big\}, \\
    \mathcal R_{\text{intra}} &= \begin{cases}
        \big\{ (\mathbf h_i^\tau \xrightarrow{{\text{past}}} \mathbf h_j^\tau) |\, i - p < j < i, \tau \in \{ t,a,v \} \big\} \\
        \big\{ (\mathbf h_i^\tau \xrightarrow{\text{future}} \mathbf h_j^\tau) |\, i < j < i + f, \tau \in \{ t,a,v \} \big\}
    \end{cases}
\end{align}

\textbf{Edges.}$\quad$The edge $(\mathbf h_i^\tau, \mathbf h_j^\nu, r_{ij}) \in \mathcal E_{\mathcal G}; \tau,\nu \in \{t,a,v\}$ represents the interaction between $\mathbf h_i^\tau$ and $\mathbf h_j^\nu$ with the relation type $r_{ij} \in \mathcal R_{\mathcal G}$. Following \cite{skianis-etal-2018-fusing}, we utilize the angular similarity to represent the edge weight between two nodes: $\mathbf A_{ij} = 1 - \arccos (\mathrm{sim}(\mathbf h_i^\tau, \mathbf h_j^\nu))/\pi$, where $\text{sim}(\cdot)$ is cosine similarity function.

\subsubsection{Inception Graph Module}

The range of emotional influence varies between utterances across different conversations. In contexts with significant fluctuations, shorter interaction distances exert a stronger emotional impact, whereas in more stable conversations, longer distances also contribute to the target utterance's emotional tone. Consequently, determining the optimal sliding window $[p, f]$ for graph construction poses a significant challenge. Drawing inspiration from the usage of multiple filter sizes as proposed in \cite{Szegedy_2015_CVPR}, we design multiple graph structures corresponding to $n$ distinct window slides $\mathcal P = \{ [p_1, f_1], \ldots, [p_n, f_n] \}$. Each graph utilizes a different slide, enabling the parallel learning of multi-scale features, which are subsequently combined to form a comprehensive and rich representation. \Cref{fig:framework}A illustrates the module, while \Cref{fig:framework}B depicts the graph structure of an individual block.

\subsubsection{Graph Learning}

With the objective of leveraging the variations of heterogeneous interactions between utterances and modalities as well as the structure diversity of multiple graph blocks, we employ $k$-dimensional GNNs ($k$-GNNs) \cite{Morris2019kGNNs}. Specifically, the representation for the $i$-th utterance at layer $\ell$ $(0 < \ell \leq N_{\text{inc}})$ is inferred as follows:
\begin{equation}\label{eq:kgnns}
    \mathbf{g}_{i,(\ell)}^\tau = \frac{1}{|\mathcal N(i)|} \sum_{r \in \mathcal R} \Big(\mathbf W_0^r \mathbf{g}_{i,(\ell-1)}^\tau + \mathbf W_1^r \sum_{j \in \mathcal N_r(i)} \mathbf A_{ji} \mathbf{g}_{j,(\ell-1)}^\nu \Big),
\end{equation}
where $\mathbf{g}_{i,(0)}^\tau = \mathbf h_i^\tau$; $\mathcal N_r(i)$ is the set of the node $i$'s neighbors with the relation $r \in \mathcal R$ and $|\mathcal N(i)| = \sum_{r \in \mathcal R} |\mathcal N_r(i)|$; $\mathbf W_0^r, \mathbf W_1^r \in \mathbb R^{d_h \times d_h}$ are learnable parameters. After $\ell = N_{\text{inc}}$ iterations, we feed the output $\mathbf g_{i}^\tau = \mathbf g_{i,(N_{\text{inc}})}^\tau$ into a Graph Transformer model \cite{ijcai2021p214} to further extract rich representations. The representation is then transformed into:
\begin{equation}\label{eq:graph-trans}
    \mathbf{o}_{i}^\tau = ||_{h=1}^H \big[ \mathbf W_2 \mathbf{g}_{i}^\tau + \sum_{j\in \mathcal N(i)} \alpha_{ij}^\tau \mathbf W_3 \mathbf{g}_{j}^\tau \big],
\end{equation}
where $\mathbf{W}_2, \mathbf{W}_3 \in \mathbb{R}^{d_h \times d_h}$ are learnable parameters, and $||_{h=1}^H$ represents the concatenation of outputs from $H$ attention heads. The attention coefficient $\alpha_{ij}^\tau$ is determined by:
\begin{equation}
    \alpha_{ij}^\tau = \mathrm{softmax}\bigg( \frac{(\mathbf W_4 \mathbf{g}_{i}^\tau)^\top (\mathbf W_5 \mathbf{g}_{j}^\tau)}{\sqrt{d_h}} \bigg),
\end{equation}
where $\mathbf W_4, \mathbf W_5 \in \mathbb R^{d_h \times d_h}$ are learnable parameters. Finally, we aggregate the representation across every branch of the module, to create a unified representation that capable of capturing multi-scale interactions among modalities and utterances. As a result, we obtain new representation vectors:
\begin{equation}
    \mathbf P^\tau = [\mathbf p_1^\tau, \mathbf p_2^\tau, \ldots, \mathbf p_L^\tau], \quad \tau \in \{t,a,v\},
\end{equation}
where $\mathbf p_i^\tau = \frac{1}{n}\sum_{j=1}^n \big[\mathbf{o}_{i}^\tau\big]_j$ and $\mathbf p_i^\tau \in \mathbb R^{d_h}$.

\subsection{Hypergraph Module (HM)}

\subsubsection{Graph Construction}

We construct a hypergraph $\mathcal H = (\mathcal V_{\mathcal H}, \mathcal E_{\mathcal H}, \omega)$ from a sequence of $L$ utterances. Similar to the ones in $\mathcal G$, each node $v \in \mathcal V_{\mathcal H}$ $(|\mathcal V_{\mathcal H}| = 3L)$ represents a unimodal utterance. We intialize the node embeddings $\{ \mathbf q_{i,(0)}^t, \mathbf q_{i,(0)}^a, \mathbf q_{i,(0)}^v \}$ with encoded representations $\{ \mathbf h_i^t, \mathbf h_i^a, \mathbf h_i^v \}$ respectively. Different from $\mathcal G$, every hyperedges $e \in \mathcal E_{\mathcal H}$ $(|\mathcal E_{\mathcal H}| = 3+L)$ are designed to capture the combined effect of modalities and conversational context, connecting every nodes within the same modality and across different modalities in a same utterance. In this fashion, the constructed hypergraph is able to capture high-order and multivariate messages that are beyond pairwise formulation. Additionally, we introduce learnable edge weight $\omega(e)$ for every hyperedge $e$, enhancing the representation of complex multivariate relationships.

\subsubsection{Graph Learning}

We employ hypergraph convolution operation \cite{BAI2021107637} to propagate multivariate embeddings. Mathematically,
\begin{equation}
    \mathbf Q^{(l)} = \sigma (\mathbf D^{-1} \mathbf H \mathbf W_e \mathbf B^{-1} \mathbf H^\top \mathbf Q^{(l-1)} \boldsymbol\Theta),
\end{equation}
where $\mathbf Q^{(l)} = \big\{ \mathbf q_{i,(l)}^\tau |\, i \in [1, L], \tau \in \{t,a,v\} \big\} \in \mathbb R^{|\mathcal V_{\mathcal H}| \times d_h}$ is the input at layer $l$. $\sigma$ is a non-linear activation function. $\mathbf H \in \{ 0, 1 \}^{|\mathcal V_{\mathcal H}| \times |\mathcal E_{\mathcal H}|}$ represents the incidence matrix, $\mathbf W_e = \mathrm{diag}(\omega(e_1),\ldots, \omega(e_{|\mathcal E_{\mathcal H}|}))$ is the learnable diagonal hyperedge weight matrix, and $\mathbf D \in \mathbb R^{|\mathcal V_{\mathcal H}| \times |\mathcal V_{\mathcal H}|}$ and $\mathbf B \in \mathbb R^{|\mathcal E_{\mathcal H}| \times |\mathcal E_{\mathcal H}|}$ are the node degree matrices and hyperedge degree matrix, respectively. After completing $N_{\text{hyp}}$ iterations, the final iteration's outputs are obtained as the multivariate representations: 
\begin{equation}
    \mathbf Q^\tau = [\mathbf q_1^\tau, \mathbf q_2^\tau, \ldots, \mathbf q_L^\tau],\quad \tau \in \{t,a,v\},
\end{equation}
in which $\mathbf q_i^\tau = \mathbf q_{i,(N_{\text{hyp}})}^\tau$.

\subsection{Fusion Module and Classifier}

After utilizing the two mentioned modules, we combine their outputs by concatenating them to form the final feature representation $\mathbf f^\tau_i = \mathbf W_6 [\mathbf p_i^\tau \, || \, \mathbf q_i^\tau] + \mathbf b_6$, where $\mathbf W_6 \in \mathbb R^{d_a \times 2d_h}$ and $\mathbf b_6 \in \mathbb R^{d_a}$ are learnable parameters. Given that the textual modality carries more sentiment information \cite{nguyen-etal-2023-conversation}, we propose a cross-modal attention mechanism to align the other two modalities with the textual features, resulting in fused representations for the text-vision and text-audio modalities. Specifically, we define the cross-modal attention mechanism as follows:
\begin{equation}
    \mathrm{CA}_i^{\tau \rightarrow t} = \mathrm{Softmax}\bigg( \frac{(\mathbf W_Q \mathbf f^\tau_i)^\top (\mathbf W_K \mathbf f^t_i)}{\sqrt{d_h}} \bigg) \mathbf W_V \mathbf f_i^t,
\end{equation}
where $\tau \in \{ a,v\}$. $\mathbf W_Q, \mathbf W_K, \mathbf W_V \in \mathbb R^{d_a \times d_a}$ are the query, key, and value weights, respectively. Then, the two fused results were added to the textual features to form a new textual representation after the crossmodal attention: 
\begin{equation}
    \hat{\mathbf f}_i^t = \mathbf f_i^t + \mathrm{CA}_i^{v \rightarrow t} + \mathrm{CA}_i^{a \rightarrow t}
\end{equation}

We then aggregate the feature representations of the three modalities to create a unified, low-dimensional feature representation using a fully-connected layer and ReLU function:
\begin{equation}
    \mathbf z_i = \mathrm{ReLU}(\mathbf W_z [\hat{\mathbf f}_i^t \, || \, \mathbf f_i^a \, || \, \mathbf f_i^v] + \mathbf b_z) \in \mathbb R^{d_z}
\end{equation}

Finally, $\mathbf z_i$ is then fed to a classifier, which is a fully-connected layer, to predict the emotion label $y_i$ for the utterance $u_i$:
\begin{align}
    \mathbf p_i &= \mathrm{Softmax}(\mathbf W_7 \mathbf z_i + \mathbf b_7) \\
    \hat{\mathbf y}_i &= \arg\max(\mathbf p_i)
\end{align}
where $\mathbf W_7, \mathbf b_7$ are trainable parameters.

\subsection{Training Objectives}

To address the challenge of classifying minority classes during training, it is crucial to mitigate the impact of imbalanced class distributions. Inspired from the approach in \cite{Cui_2019_CVPR}, we introduce a re-weighting strategy that uses the effective number of samples for each class to adjust the loss, resulting in a class-balanced loss. Specifically, for a sample from class $c_i$ with $n_i$ total samples, a weighting factor $w_c(i) = {(1-\beta)} / {(1-\beta^{n_i})}$ is applied to the loss function, where $\beta \in [0,1)$ is a hyperparameter. Given a batch of $N$ dialogues, where the $i$-th dialogue contains $L_i$ utterances, the class-balanced (CB) training objectives are defined as follows: 

\textbf{Focal Contrastive Loss.}\quad To address the challenge of classifying minority classes, we introduce a novel loss function called Class-Balanced Focal Contrastive (CBFC) loss, which extends the focal contrastive loss \cite{NEURIPS2021_fa14d4fe} by incorporating a class-weight term. This loss aligns pairs with the same emotional labels and maximizes inter-class distances by pushing apart pairs with different labels. The CBFC loss is formulated as follows:
\begin{equation}
    \mathcal L_{\text{CBFC}} = - \frac{1}{\sum_{i=1}^N L_i} \sum_{i=1}^N \sum_{j=1}^{L_i} \frac{w_{c}(j)}{|\mathcal P_{i,j}|} \sum_{\mathbf z_{i,k} \in \mathcal P_{i,j}} (1 - t_{j,k}^{(i)}) \log t_{j,k}^{(i)},
\end{equation}
where $t_{j,k}^{(i)} = \frac{\exp(\mathbf z_{i,j}^\top \mathbf z_{i,k} / \tau)}{\sum_{\mathbf z_{i,s} \in \mathcal A_{i,j}} \exp(\mathbf z_{i,j}^\top \mathbf z_{i,s} / \tau) }$, in which $\mathcal P_{i,j}, \mathcal A_{i,j}$ denote the anchor's positive and full pair sets.

\textbf{Cross-Entropy Loss.}\quad We adopt a weighted Cross-Entropy (CE) loss to measure the difference between predicted probabilities and true labels:
\begin{equation}
    \mathcal L_{\text{CBCE}} = -\frac{1}{\sum_{i=1}^N L_i} \sum_{i=1}^N \sum_{j=1}^{L_i} w_{c}(j) \sum_{c=1}^{|C|} \mathbf y_{j}^c \log \mathbf p_{j}^c,
\end{equation}
where $\mathbf y_{j}^c$ is the one-hot vector of the true label.

\textbf{Full Loss.}\quad We linearly combine focal contrastive loss and Cross-entropy loss as follows:
\begin{equation}
    \mathcal L = \mathcal L_{\text{CBCE}} + \mu \mathcal L_{\text{CBFC}},
\end{equation}
where $\mu \in (0, 1]$ is a tunable hyperparameter.

\section{Experiments and Analysis}\label{sec:experiments}

\subsection{Dataset}
We conducted experiments on two multimodal datasets: IEMOCAP \cite{Busso2008} and MELD \cite{poria-etal-2019-meld}. The IEMOCAP dataset consists of 12 hours of two-way conversations involving 10 speakers, comprising a total of 7,433 utterances and 151 dialogues, categorized into six emotion classes: happy, sad, neutral, angry, excited, and frustrated. The MELD dataset includes 1,433 conversations and 13,708 utterances, each labeled with one of seven emotion categories: angry, disgusted, fearful, happy, sad, surprised, and neutral. To ensure a fair comparison, we utilized the predefined train/validation/test splits provided by each dataset. As IEMOCAP lacks a validation set, we followed the split used in recent work \cite{nguyen-etal-2023-conversation} for training and validating all methods.

\subsection{Experimental Setups}

For both datasets, we use the Adam optimizer \cite{kingma2017adammethodstochasticoptimization} with a learning rate of 0.0004 over 40 epochs. The number of layers in IGM and HM for both datasets is set to 2 and 4, respectively. We set $\beta=0.999$ and $\mu=0.8$ across both datasets. The IGM architecture comprises 3 GNN branches, with window sizes set to [(10, 9), (5, 3), (3, 2)] for IEMOCAP and [(11, 11), (7, 4), (6, 4)] for MELD. Hyperparameters, including window sizes and the number of layers in each module, are set using the validation set. All reported results represent the mean of five independent runs.

\subsection{Experimental Results}

\begin{table}[t]
    \centering
    \caption{Comparison with prior SOTA methods on IEMOCAP and MELD.}\label{tab:main}
    \begin{tabular}{c|lc|cc}
        \toprule
        & Method & Network & Acc ($\%$) & w-F1 ($\%$) \\
        \midrule
        \verticalrow{6}{IEMOCAP} 

        & DialogueGCN \cite{ghosal-etal-2019-dialoguegcn} & GNN-based 
        & 55.29 & 55.16 \\
        & DialogueRNN \cite{Majumder2019dialrnn} & Non-GNN & 57.22 & 55.29 \\
        & ICON \cite{hazarika-etal-2018-icon} & Non-GNN & 63.10 & 63.8 \\
        & COGMEN \cite{joshi-etal-2022-cogmen} & GNN-based & 64.02 & 63.78 \\
        & CORECT \cite{nguyen-etal-2023-conversation} & GNN-based & 66.20 & 66.39 \\
        \cmidrule{2-5}
        & ConxGNN \textit{(ours)} & GNN-based & \textbf{68.52} & \textbf{68.64} \\
        \midrule
        \verticalrow{5}{MELD} 
        & DialogueGCN \cite{ghosal-etal-2019-dialoguegcn} & GNN-based 
        & 42.75 & 41.67 \\
        & DialogueRNN \cite{Majumder2019dialrnn} & Non-GNN 
        & 61.88 & 61.63 \\
        & MM-DFN \cite{9747397} & GNN-based 
        & 66.09 & 64.16 \\ 
        & M$^3$Net \cite{10203083} & GNN-based
        & 65.75 & 65.00 \\
        \cmidrule{2-5}
        & ConxGNN \textit{(ours)} & GNN-based & \textbf{66.28} & \textbf{65.69} \\
        \bottomrule
    \end{tabular}
\end{table}

\Cref{tab:main} presents a performance comparison between our proposed method and other SOTA approaches. The results demonstrate that ConxGNN achieves superior performance across both datasets. Specifically, ConxGNN surpasses the previous best method, CORECT \cite{nguyen-etal-2023-conversation}, by $2.32\%$ in accuracy and $2.25\%$ in weighted-F1 score on the IEMOCAP dataset. On the MELD dataset, our model shows slightly improvements of $0.19\%$ in accuracy and $0.69\%$ in weighted-F1 score compared to MM-DFN \cite{9747397} and M$^3$Net \cite{Cui_2019_CVPR}, respectively. These findings empirically validate the effectiveness of our proposed architecture.

\subsection{Ablation Study}

\subsubsection{Components Analysis}

We conduct ablation study to evaluate the contribution of each module within our framework. \Cref{tab:components} represents the model's performance when specific components are removed. Of the four modules, we can see that IGM has the greatest impact, as its removal leads to a significant performance decline across both datasets, with a drop in (accuracy, weighted-F1) of $(27.8\%, 42.96\%)$ on IEMOCAP and $(15.44\%, 25.48\%)$ on MELD. The second key module, HM, also plays a critical role, especially on IEMOCAP, where its absence results in approximately a $4.5\%$ reduction in performance, though its effect on MELD is minimal, causing around $1\%$ degradation. The removal of other components, such as the cross-modal attention mechanism and the re-weighting scheme, also results in slight performance reductions. These findings collectively confirm the importance and effectiveness of each component in our architecture.

\subsubsection{Impact of Multi-scale Extractor} 
To highlight the significance of the IGM, we conduct an ablation study by varying the number of inception graph blocks/branches within the module. \Cref{tab:multiscale} presents the best average results for each number of blocks. In the 2-block analysis, we explore different combinations of three sliding windows to evaluate performance. The results are fairly consistent for the same number of blocks. Additionally, performance improves steadily as more blocks are added, with an approximate increase of $1\%$ per block. Compared to the single-scale approach (i.e., a single block), our multi-scale strategy leads to notable performance gains, with accuracy increasing by $3.15\%$ on IEMOCAP and $2.83\%$ on MELD, and weighted F1 improving by $3.09\%$ on IEMOCAP and $2.83\%$ on MELD. These findings underscore the importance of the proposed IGM, which captures multi-scale interactions between modalities and utterances.

\begin{table}[t]
    \centering
    \caption{Performance with different strategies.} \label{tab:components}
    \begin{tabular}{lcc|cc}
        \toprule
        \multirow{2}{*}{Method} & \multicolumn{2}{c|}{IEMOCAP} & \multicolumn{2}{c}{MELD} \\
        \cmidrule{2-3} \cmidrule{4-5}
        & Acc $(\%)$ & w-F1 $(\%)$ & Acc $(\%)$ & w-F1 $(\%)$ \\
        \midrule
        ConxGNN & \textbf{68.52} & \textbf{68.64} & \textbf{66.28} & \textbf{65.69} \\
        $-$ w/o IGM & 38.48 & 25.68 & 50.84 & 40.21 \\
        $-$ w/o HM & 64.06 & 63.92 & 65.11 & 64.87 \\
        $-$ w/o crossmodal & 64.21 & 64.31 & 66.15 & 65.69 \\
        $-$ w/o re-weight & 63.13 & 63.90 & 65.30 & 65.10 \\
        \bottomrule
    \end{tabular}
\end{table}

\begin{table}[t]
    \centering
    \caption{Performance with different number of blocks.} \label{tab:multiscale}
    \begin{tabular}{ccc|cc}
        \toprule
        \multirow{2}{*}{\# Blocks} & \multicolumn{2}{c|}{IEMOCAP} & \multicolumn{2}{c}{MELD} \\
        \cmidrule{2-3} \cmidrule{4-5}
        & Acc $(\%)$ & w-F1 $(\%)$ & Acc $(\%)$ & w-F1 $(\%)$ \\
        \midrule
        \multirow{3}{*}{1} & 65.27 & 65.34 & 64.36 & 62.61 \\
        & 65.29 & 65.31 & 64.27 & 62.65 \\
        & \textbf{65.37} & \textbf{65.55} & \textbf{64.70} & \textbf{62.86} \\
        \midrule
        \multirow{3}{*}{2} & 66.30 & 66.64 & 65.34 & 63.49 \\
        & 66.02 & 65.88 & 65.40 & 63.44 \\
        & \textbf{66.74} & \textbf{66.91} & \textbf{65.81} & \textbf{63.88} \\
        \midrule
        3 & \textbf{68.52} & \textbf{68.64} & \textbf{66.28} & \textbf{65.69} \\
        \bottomrule
    \end{tabular}
\end{table}

\section{Conclusion}\label{sec:conclusion}

We propose ConxGNN, a novel framework specifically designed for contextual modeling in conversations for the ERC task. ConxGNN is composed of two primary modules: the IGM, which extracts multi-scale relationships using varying interactive window sizes, and HM, which captures the multivariate relationships among utterances and modalities. These modules operate in parallel, and their outputs are combined using an attention mechanism, resulting in contextually enriched information. Additionally, ConxGNN addresses the issue of class imbalance by incorporating a re-weighting scheme into the loss functions. Experimental results on the IEMOCAP and MELD datasets demonstrate that our approach achieves SOTA performance, highlighting its efficacy and advantages.

\clearpage
\bibliographystyle{IEEEbib}
\bibliography{main}

\begin{thebibliography}{10}

\bibitem{hu-etal-2021-dialoguecrn}
Hu et~al.,
\newblock ``{D}ialogue{CRN}: Contextual reasoning networks for emotion
  recognition in conversations,''
\newblock in {\em Proceedings of the 59th Annual Meeting of the Association for
  Computational Linguistics and the 11th International Joint Conference on
  Natural Language Processing (Volume 1: Long Papers)}, Zong et~al., Eds.,
  Online, Aug. 2021, pp. 7042--7052, Association for Computational Linguistics.

\bibitem{10446496}
Chandola et~al.,
\newblock ``Serc-gcn: Speech emotion recognition in conversation using graph
  convolutional networks,''
\newblock in {\em ICASSP 2024 - 2024 IEEE International Conference on
  Acoustics, Speech and Signal Processing (ICASSP)}, 2024, pp. 76--80.

\bibitem{hazarika-etal-2018-icon}
Hazarika et~al.,
\newblock ``{ICON}: Interactive conversational memory network for multimodal
  emotion detection,''
\newblock in {\em Proceedings of the 2018 Conference on Empirical Methods in
  Natural Language Processing}, Riloff et~al., Eds., Brussels, Belgium,
  Oct.-Nov. 2018, pp. 2594--2604, Association for Computational Linguistics.

\bibitem{hazarika-etal-2018-conversational}
Hazarika et~al.,
\newblock ``Conversational memory network for emotion recognition in dyadic
  dialogue videos,''
\newblock in {\em Proceedings of the 2018 Conference of the North {A}merican
  Chapter of the Association for Computational Linguistics: Human Language
  Technologies, Volume 1 (Long Papers)}, Walker et~al., Eds., New Orleans,
  Louisiana, June 2018, pp. 2122--2132, Association for Computational
  Linguistics.

\bibitem{Majumder2019dialrnn}
Majumder et~al.,
\newblock ``Dialoguernn: An attentive rnn for emotion detection in
  conversations,''
\newblock {\em Proceedings of the AAAI Conference on Artificial Intelligence},
  vol. 33, no. 01, pp. 6818--6825, Jul. 2019.

\bibitem{10.1145/3664647.3681648}
Nguyen et~al.,
\newblock ``Ada2i: Enhancing modality balance for multimodal conversational
  emotion recognition,''
\newblock in {\em Proceedings of the 32nd ACM International Conference on
  Multimedia}, New York, NY, USA, 2024, MM '24, p. 9330–9339, Association for
  Computing Machinery.

\bibitem{ghosal-etal-2019-dialoguegcn}
Ghosal et~al.,
\newblock ``{D}ialogue{GCN}: A graph convolutional neural network for emotion
  recognition in conversation,''
\newblock in {\em Proceedings of the 2019 Conference on Empirical Methods in
  Natural Language Processing and the 9th International Joint Conference on
  Natural Language Processing (EMNLP-IJCNLP)}, Inui et~al., Eds., Hong Kong,
  China, Nov. 2019, pp. 154--164, Association for Computational Linguistics.

\bibitem{ishiwatari-etal-2020-relation}
Ishiwatari et~al.,
\newblock ``Relation-aware graph attention networks with relational position
  encodings for emotion recognition in conversations,''
\newblock in {\em Proceedings of the 2020 Conference on Empirical Methods in
  Natural Language Processing (EMNLP)}, Webber et~al., Eds., Online, Nov. 2020,
  pp. 7360--7370, Association for Computational Linguistics.

\bibitem{shen-etal-2021-directed}
Shen et~al.,
\newblock ``Directed acyclic graph network for conversational emotion
  recognition,''
\newblock in {\em Proceedings of the 59th Annual Meeting of the Association for
  Computational Linguistics and the 11th International Joint Conference on
  Natural Language Processing (Volume 1: Long Papers)}, Zong et~al., Eds.,
  Online, Aug. 2021, pp. 1551--1560, Association for Computational Linguistics.

\bibitem{nguyen-etal-2023-conversation}
Nguyen et~al.,
\newblock ``Conversation understanding using relational temporal graph neural
  networks with auxiliary cross-modality interaction,''
\newblock in {\em Proceedings of the 2023 Conference on Empirical Methods in
  Natural Language Processing}, Bouamor et~al., Eds., Singapore, Dec. 2023, pp.
  15154--15167, Association for Computational Linguistics.

\bibitem{10.1145/3593583}
Zhang et~al.,
\newblock ``M3gat: A multi-modal, multi-task interactive graph attention
  network for conversational sentiment analysis and emotion recognition,''
\newblock {\em ACM Trans. Inf. Syst.}, vol. 42, no. 1, Aug. 2023.

\bibitem{li-etal-2023-joyful}
Li et~al.,
\newblock ``Joyful: Joint modality fusion and graph contrastive learning for
  multimoda emotion recognition,''
\newblock in {\em Proceedings of the 2023 Conference on Empirical Methods in
  Natural Language Processing}, Bouamor et~al., Eds., Singapore, Dec. 2023, pp.
  16051--16069, Association for Computational Linguistics.

\bibitem{10.1145/3664647.3681633}
Yi et~al.,
\newblock ``Multimodal fusion via hypergraph autoencoder and contrastive
  learning for emotion recognition in conversation,''
\newblock in {\em Proceedings of the 32nd ACM International Conference on
  Multimedia}, New York, NY, USA, 2024, MM '24, p. 4341–4348, Association for
  Computing Machinery.

\bibitem{10095097}
Zhang et~al.,
\newblock ``Knowledge-aware graph convolutional network with utterance-specific
  window search for emotion recognition in conversations,''
\newblock in {\em ICASSP 2023 - 2023 IEEE International Conference on
  Acoustics, Speech and Signal Processing (ICASSP)}, 2023, pp. 1--5.

\bibitem{Szegedy_2015_CVPR}
Szegedy et~al.,
\newblock ``Going deeper with convolutions,''
\newblock in {\em Proceedings of the IEEE Conference on Computer Vision and
  Pattern Recognition (CVPR)}, June 2015.

\bibitem{NIPS2017_3f5ee243}
Vaswani et~al.,
\newblock ``Attention is all you need,''
\newblock in {\em Advances in Neural Information Processing Systems}, Guyon
  et~al., Eds. 2017, vol.~30, Curran Associates, Inc.

\bibitem{skianis-etal-2018-fusing}
Skianis et~al.,
\newblock ``Fusing document, collection and label graph-based representations
  with word embeddings for text classification,''
\newblock in {\em Proceedings of the Twelfth Workshop on Graph-Based Methods
  for Natural Language Processing ({T}ext{G}raphs-12)}, Glava{\v{s}} et~al.,
  Eds., New Orleans, Louisiana, USA, June 2018, pp. 49--58, Association for
  Computational Linguistics.

\bibitem{Morris2019kGNNs}
Morris et~al.,
\newblock ``Weisfeiler and leman go neural: Higher-order graph neural
  networks,''
\newblock {\em Proceedings of the AAAI Conference on Artificial Intelligence},
  vol. 33, no. 01, pp. 4602--4609, Jul. 2019.

\bibitem{ijcai2021p214}
Shi et~al.,
\newblock ``Masked label prediction: Unified message passing model for
  semi-supervised classification,''
\newblock in {\em Proceedings of the Thirtieth International Joint Conference
  on Artificial Intelligence, {IJCAI-21}}, Zhi-Hua Zhou, Ed. 8 2021, pp.
  1548--1554, International Joint Conferences on Artificial Intelligence
  Organization,
\newblock Main Track.

\bibitem{BAI2021107637}
Bai et~al.,
\newblock ``Hypergraph convolution and hypergraph attention,''
\newblock {\em Pattern Recognition}, vol. 110, pp. 107637, 2021.

\bibitem{Cui_2019_CVPR}
Cui et~al.,
\newblock ``Class-balanced loss based on effective number of samples,''
\newblock in {\em Proceedings of the IEEE/CVF Conference on Computer Vision and
  Pattern Recognition (CVPR)}, June 2019.

\bibitem{NEURIPS2021_fa14d4fe}
Zhang et~al.,
\newblock ``Unleashing the power of contrastive self-supervised visual models
  via contrast-regularized fine-tuning,''
\newblock in {\em Advances in Neural Information Processing Systems}, Ranzato
  et~al., Eds. 2021, vol.~34, pp. 29848--29860, Curran Associates, Inc.

\bibitem{Busso2008}
Busso et~al.,
\newblock ``Iemocap: interactive emotional dyadic motion capture database,''
\newblock {\em Language Resources and Evaluation}, vol. 42, no. 4, pp.
  335--359, Dec 2008.

\bibitem{poria-etal-2019-meld}
Poria et~al.,
\newblock ``{MELD}: A multimodal multi-party dataset for emotion recognition in
  conversations,''
\newblock in {\em Proceedings of the 57th Annual Meeting of the Association for
  Computational Linguistics}, Korhonen et~al., Eds., Florence, Italy, July
  2019, pp. 527--536, Association for Computational Linguistics.

\bibitem{kingma2017adammethodstochasticoptimization}
Diederik~P. Kingma and Jimmy Ba,
\newblock ``Adam: A method for stochastic optimization,'' 2017.

\bibitem{joshi-etal-2022-cogmen}
Joshi et~al.,
\newblock ``{COGMEN}: {CO}ntextualized {GNN} based multimodal emotion
  recognitio{N},''
\newblock in {\em Proceedings of the 2022 Conference of the North American
  Chapter of the Association for Computational Linguistics: Human Language
  Technologies}, Carpuat et~al., Eds., Seattle, United States, July 2022, pp.
  4148--4164, Association for Computational Linguistics.

\bibitem{9747397}
Hu et~al.,
\newblock ``Mm-dfn: Multimodal dynamic fusion network for emotion recognition
  in conversations,''
\newblock in {\em ICASSP 2022 - 2022 IEEE International Conference on
  Acoustics, Speech and Signal Processing (ICASSP)}, 2022, pp. 7037--7041.

\bibitem{10203083}
Chen et~al.,
\newblock ``Multivariate, multi-frequency and multimodal: Rethinking graph
  neural networks for emotion recognition in conversation,''
\newblock in {\em 2023 IEEE/CVF Conference on Computer Vision and Pattern
  Recognition (CVPR)}, 2023, pp. 10761--10770.

\end{thebibliography}
\end{document}